# Making Deep Heatmaps Robust to Partial Occlusions for 3D Object Pose Estimation

Markus Oberweger[1][0000−0003−4247−2818], Mahdi Rad[1][0000−0002−4011−4729], and Vincent Lepetit[2,1][0000−0001−9985−4433]

[1] Institute for Computer Graphics and Vision, Graz University of Technology, Graz, Austria
[2] Laboratoire Bordelais de Recherche en Informatique, Université de Bordeaux, Bordeaux, France
{oberweger,rad,lepetit}@icg.tugraz.at

**Abstract.** We introduce a novel method for robust and accurate 3D object pose estimation from a single color image under large occlusions. Following recent approaches, we first predict the 2D projections of 3D points related to the target object and then compute the 3D pose from these correspondences using a geometric method. Unfortunately, as the results of our experiments show, predicting these 2D projections using a regular CNN or a Convolutional Pose Machine is highly sensitive to partial occlusions, even when these methods are trained with partially occluded examples. Our solution is to predict heatmaps from multiple small patches independently and to accumulate the results to obtain accurate and robust predictions. Training subsequently becomes challenging because patches with similar appearances but different positions on the object correspond to different heatmaps. However, we provide a simple yet effective solution to deal with such ambiguities. We show that our approach outperforms existing methods on two challenging datasets: The Occluded LineMOD dataset and the YCB-Video dataset, both exhibiting cluttered scenes with highly occluded objects.

## 1 Introduction

3D object pose estimation from images is an old but currently highly researched topic, mostly due to the advent of Deep Learning-based approaches and the possibility of using large datasets for training such methods. 3D object pose estimation from RGB-D already has provided compelling results [1–4], and the accuracy of methods that only require RGB images recently led to huge progress in the field [5–7, 2–4, 8]. In particular, one way to obtain an accurate pose is to rely on a Deep Network to initially predict the 2D projections of some chosen 3D points and then compute the 3D pose of the object using a P$n$P method [9]. Such an approach has been shown to be more accurate than the approach of directly predicting the pose used in [5–7], and, therefore, we used the former approach in the research described in this paper.

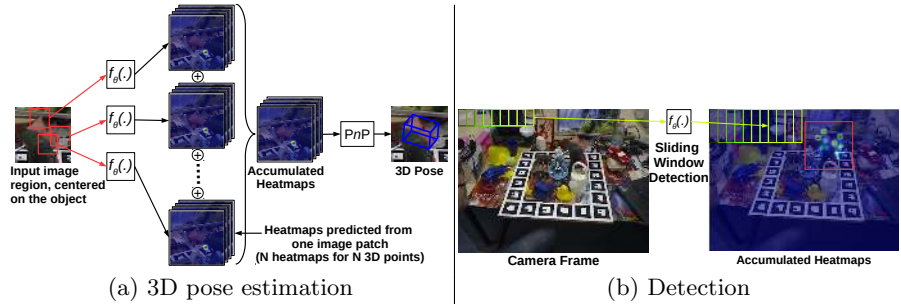

(a) 3D pose estimation            (b) Detection

Fig. 1: Overview of our method. (a) Given an image region centered on the target object, we sample image patches from which we predict heatmaps for the 2D projections of the corners of the object's 3D bounding box. This prediction is done by a Deep Network $f_\theta(\cdot)$. We aggregate the heatmaps and extract the global maxima for each heatmap, from which we compute the 3D object pose using a P$n$P algorithm. *We show that $f_\theta(\cdot)$ can be trained simply and efficiently despite the ambiguities that may arise when using small patches as input.* (b) To obtain the image region centered on the object, we apply the predictor in a sliding window fashion and accumulate the heatmaps for the full camera frame. We keep the image region with the largest values after accumulation.

However, while Deep Learning methods allow researchers to predict the pose of fully visible objects, they suffer significantly from occlusions, which are very common in practice: Parts of the target object can be hidden by other objects or by a hand interacting with the object. A common *ad hoc* solution is to train the network with occluded objects in the training data. As the results of our experiments presented in this paper show, the presence of large occlusions and unknown occluders still decrease the accuracy of the predicted pose.

Instead of using the entire image of the target object as input to the network, we consider image patches, as illustrated in Fig. 1, since at least some of these are not corrupted by the occluder. Using an image patch as input, our approach learns to predict heatmaps over the 2D projections of 3D points related to the target object. By combining the heatmaps predicted from many patches, we obtain an accurate 3D pose even if some patches actually lie on the occluder or the background instead of on the object.

When moving to an image patch as input, the prediction becomes multimodal. This is shown in Fig. 2: Some patches may appear on different parts of the target object but look similar. These patches are ambiguous, as they can correspond to different predictions. In such a case, we would like to predict heatmaps with multiple local maxima, one for each possible prediction. The main challenge is that the ambiguities are difficult to identify: This would require us to identify the patches that have similar appearances, from all the possible viewpoints and at all the possible positions on the object.

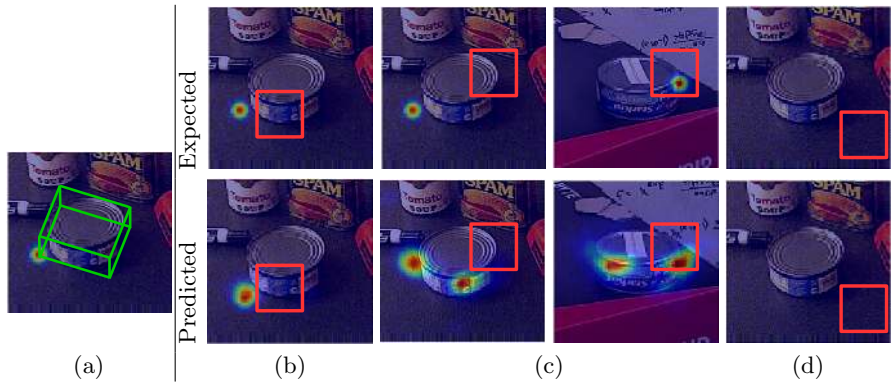

Fig. 2: Predicting heatmaps from image patches. In this example, we consider predicting the projection of the 3D corner highlighted in (a) for the *tuna fish can* object of the YCB-Video dataset [3]. The red boxes show the input patch of the predicted heatmap. (b) shows a patch from which the projection can be predicted unambiguously. (c) shows two patches that are located in two different positions on the can (notice that the can is flipped and rotated between the two images) while having a similar appearance. In presence of such patches, it is only possible to predict a distribution over the possible locations for the projection. (d) shows a patch on the background, from which we predict a uniform heatmap as it does not provide additional information. See text for details.

The authors of [10] faced a similar problem in the context of 2D object detection when aiming to localize semantic parts from feature vectors of a convolutional layer computed by a CNN. As we discuss in Section 2, the method they proposed is complex both for training and inference, and also inaccurate. The solution we propose is much simpler yet efficient: We train a network to predict heatmaps corresponding to a single solution for training image patches using a least-squares loss function. Thanks to the properties of the least-squares loss, this makes the network naturally predict the *average* of the possible heatmap solutions for a given patch. This is exactly what we want, because it is the best information we can obtain from a single patch even if the information remains ambiguous. We then follow an ensemble approach and take the average of the heatmaps predicted for many patches, which allows us to resolve the ambiguities that arise with individual patches. We finally extract the global maximum from this average as the final 2D location.

Our main contribution is, therefore, a simple method that can be used to accurately predict the 3D pose of an object under partial occlusion. We also considered applying Transfer Learning to exploit additional synthetic training data and improve performances. However, as we show, if the input to a network contains an occluder, the occlusion significantly influences the network output even when the network has been trained with occlusion examples and simply adding more training data does not help. In our case, some of the input patches used by our method will not contain occluders, and Transfer Learning becomes useful. In practice, we use the Feature Mapping described in [11], which can be used to map the image features extracted from real images to corresponding image features for synthetic images. This step is not needed for our method to outperform the state-of-the-art but allows us to provide an additional performance boost.

In the remainder of this paper, we first discuss related work, then present our approach, and finally evaluate it and compare it to the state-of-the-art methods on the Occluded LineMOD [12] and the YCB-Video [3] datasets.

## 2   Related Work

The literature on 3D object pose estimation is extremely large. After the popularity of edge-based [13] and keypoint-based methods [14] waned, Machine Learning and Deep Learning became popular in recent years for addressing this problem [5–7, 2–4, 8, 15]. Here, we will mostly focus on recent work based on RGB images. In the Evaluation section, we compare our method to recent methods [5, 7, 3, 4].

[4, 8] proposed a cascade of modules, whereby the first module is used to localize the target objects, and the second module, to regress the object surface 3D coordinates. These coordinates then are used to predict the object pose through hypotheses sampling with a pre-emptive RANSAC [9]. Most importantly, we do not directly predict 3D points but average 2D heatmaps. Predicting 3D points for corresponding 2D points seems to be much more difficult than predicting 2D points for 3D points, as discussed in [3]. Also, surface coordinates are not adapted to deal with symmetric objects. In [5] the target object was also first detected, then the 2D projections of the corners of the object's 3D bounding boxes were predicted and, finally, the 3D object pose from their 3D correspondences was estimated using a P$n$P algorithm. [7] integrated this idea into a recent object detector [16] to predict 2D projections of the corners of the 3D bounding boxes, instead of a 2D bounding box. Similarly, in [6], 2D keypoints were predicted in the form of a set of heatmaps as we do in this work. However, it uses the entire image as input and, thus, performs poorly on occlusions. It also requires training images annotated with keypoint locations, while we use virtual 3D points. In [17], 2D keypoint detection was also relied upon. The authors considered partially occluded objects for inferring the 3D object location from these keypoints. However, their inference adopted a complex model fitting and required the target objects to co-occur in near-regular configuration.

In [2], the SSD architecture [18] was extended to estimate the objects' 2D locations and 3D rotations. In a next step, the authors used these predictions together with pre-computed information to estimate the object's 3D pose. However, this required a refinement step to get an accurate pose, which was influenced by occlusions. The objects were segmented in [3], and an estimate of their 3D poses was made by predicting the translation and a quaternion for the rotation, refined by ICP. Segmenting objects makes their approach robust to occlusions to some extent, however, it requires the use of a highly complex model. In [15], object parts were considered to handle partial occlusion by predicting a set of 2D-3D correspondences from each of these parts. However, the parts had to be manually picked, and it is not clear which parts can represent objects such as those we evaluate in this paper.

As mentioned in the introduction, our method is related to that described in [10]. In the context of 2D object detection, semantic parts are localized as in [10] from neighboring feature vectors using a spatial offset map. The offset maps are accumulated in a training phase. However, they need to be able to identify which feature vectors support a semantic part from these maps, and complex statistical measures are used to identify such vectors. Our method is significantly simpler, as the mapping between the input patches and the 2D projections does not have to be established explicitly.

The authors of [19] already evaluated CNNs trained on occlusions in the context of 2D object detection and recognition and proposed modifying training to penalize large spatial filters support. This yields better performance; however, this does not fully cancel out the influence of occlusions. Some recent work also describes explicitly how to handle occlusions for 3D pose estimation when dealing with 3D or RGB-D data: Like us, [20] relied on a voting scheme to increase robustness to occlusions; [21] first segmented and identified the objects from an RGB-D image. They then performed an extensive randomized search over possible object poses by considering physical simulation of the configuration. In [22], holistic and local patches were combined for object pose estimation, using a codebook for local patches and applying a nearest-neighbor search to find similar poses, as in [23, 24]. In contrast to these methods, we use only color images.

Our method is also related to ensemble methods and, in particular, the Hough Forests [25], which are based on regression trees. Hough Forests also predict 2D locations from multiple patches and are multimodal. Multimodal prediction is easy to perform with trees, as the multiple solutions can be stored in the tree leaves. With our method, we aim to combine the ability of Hough Forests for multimodal predictions and the learning power of Deep Learning. [26] already reformulated a Hough Forest as a CNN by predicting classification and regression for patches of the input image. However, this method required to handle the detection separately, and each patch regressed a single vector, which was not multimodal and required clustering of the predicted vectors. In this paper, we show that carrying out a multimodal prediction with Deep Networks to address our problem is, in fact, simple.

## 3 Influence of Occlusions on Deep Networks

In this section, we describe how we evaluate how much a partial occlusion influences a Deep Network, whether it is a standard Convolutional Neural Network (CNN) or a Convolutional Pose Machine (CPM) [27]. Specifically, a CPM is a carefully designed CNN that predicts dense heatmaps by sequentially refining results from previous stages. The input features are concatenated to intermediary heatmaps in order to learn spatial dependencies.

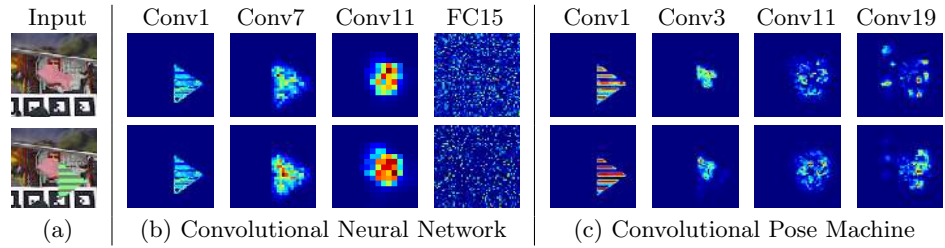

Fig. 3: Effect of occlusions on the feature maps of Deep Networks. (a) Input image without (top) and with (bottom) partial occlusion. (b-Top) Sums of absolute differences between the feature maps with and without occlusions for a CNN trained without occlusions. (b-Bottom) Same when the network is trained with occlusion examples. (c) Same for a Convolutional Pose Machine. The influence of the occlusion increases with the layers' depths, as receptive fields are larger in the deeper layers than in the first layers, even when the method is trained with occlusion examples. For more details we refer to the supplementary material.

For this experiment, depicted in Fig. 3, we use an image centered on an object as input to a network—here, the *Cat* object from the Occluded LineMOD dataset [12]. We then compare the layer activations in the absence of occlusion, and when the object is occluded by an artificial object (here, a striped triangle). We consider two networks: A standard CNN trained to predict the 2D projections of 3D points as a vector [5], and a CPM [27] with 3 stages trained to predict a heatmap for each of the same 2D projections. For the 3D points, we use the corners of the 3D bounding box of the object.

As can be seen in Fig. 3, the occlusion induces changes in the activations of all the layers of both networks. For a standard CNN, the occlusion spreads to more than 20% in the last feature map, and, beyond the first fully-connected layer, more than 45% of all activations are changed. In this case, all the predictions for the 2D projections, occluded or not, are inaccurate. A similar effect can be observed for CPMs: Here, the altered activations are more specifically localized to the occluded region due to the convolutions, with more than 29% of the activations changed in the last feature map. In this case, the predictions of the 2D projections are inaccurate when the 3D points are occluded. When the 3D points are not occluded, the predicted projections are sometimes correct, because the influence of the occluder spreads less with a CPM than with a standard CNN.

## 4   Minimizing the Effect of Occlusions

In this section, we first describe our training procedure given an input image region centered on the object, then the run-time inference of the pose. Finally, we explain how we identify the input image region in practice.

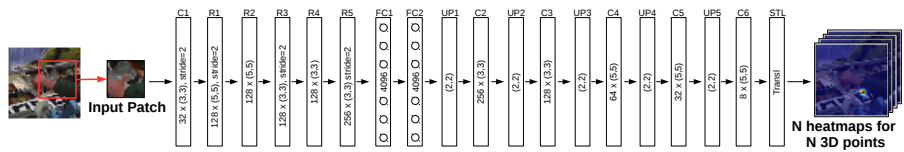

Fig. 4: Network architecture for $f_\theta(\cdot)$. C denotes a convolutional layer with the number of filters and the filter size inscribed; FC, a fully-connected layer with the number of neurons; UP, an unpooling layer [28]; R, a residual module [29] with the number of filters and filter size; and STL, a Spatial Transformation Layer [30] used for translation. All layers have ReLU activations, and the output of the last layer is linear.

## 4.1 Training

Datasets for 3D pose estimation typically provide training images annotated with the objects' poses and the 3D models of the objects. From this data, we generate our training set $\{(I^{(i)}, \{\mathbf{p}_j^{(i)}\}_j, M^{(i)})\}_i$, where $I^{(i)}$ is the $i$-th training image; $\mathbf{p}_j^{(i)}$, the 2D projection of the $j$-th 3D corner; and $M^{(i)}$, the 2D mask of the object in image $I^{(i)}$. This mask can be obtained by projecting the 3D object model into the image using the object's pose.

**The Unambiguous Case** Let us first ignore the fact that some image patches can be ambiguous and that the learning problem is actually multimodal. We train a network $f_\theta(\cdot)$ to predict a heatmap for each projection $\mathbf{p}_j$. The architecture we use for this network is shown in Fig. 4. $f_\theta(\cdot)$ takes an input patch of size $32 \times 32px$, and predicts a set of heatmaps of size $128 \times 128px$, and we train it by minimizing:

$$\min_\theta \sum_i \sum_{u,v} \|\mathcal{H}^{(i)} - \mathrm{Transl}(f_\theta(\mathcal{P}(I^{(i)}, u, v)), -u, -v)\|^2 \,, \tag{1}$$

where:

- $\mathcal{P}(I^{(i)}, u, v)$ is an image patch centered on location $(u, v)$ in image $I^{(i)}$;
- $\mathcal{H}^{(i)}$ is the set of expected heatmaps for $\mathcal{P}(I^{(i)}, u, v)$. It contains one heatmap for each 2D projection $\mathbf{p}_j^{(i)}$. We describe how $\mathcal{H}^{(i)}$ is defined in detail below;
- $f_\theta(\mathcal{P})$ returns a set of heatmaps, one for each 2D projection $\mathbf{p}_j^{(i)}$.
- $\mathrm{Transl}(H, -u, -v)$ translates the predicted heatmaps $H$ by $(-u, -v)$. $f_\theta(\cdot)$ learns to predict the heatmaps with respect to the patch center $(u, v)$, and this translation is required to correctly align the predicted heatmaps together. Such a translation can be efficiently implemented using a Spatial Transformation Layer [30], which makes the network trainable end-to-end.

The sum $\sum_{(u,v)}$ is over 2D locations randomly sampled from the image. The heatmaps in $\mathcal{H}^{(i)}$ are defined as a Gaussian distribution with a small standard deviation (we use $\sigma = 4px$ in practice) and centered on the expected 2D projections $\mathbf{p}_j^{(i)}$ when patch $\mathcal{P}(I^{(i)}, u, v)$ overlaps the object mask $M^{(i)}$. The top row of Fig. 2 shows examples of such heatmaps.

When the patch does not overlap the object mask, the heatmaps in $\mathcal{H}^{(i)}$ are defined as a uniform distribution of value $\frac{1}{W \cdot H}$, where $W \times H$ is the heatmap's resolution, since there is no information in the patch to predict the 2D projections. In addition, we use patches sampled from the ImageNet dataset [31] and train the network to predict uniform heatmaps as well for these patches. Considering these patches (outside the object's mask or from ImageNet) during training allows us to correctly handle patches appearing in the background or on the occluders and significantly reduces the number of false positives observed at run-time.

**The Multimodal Case** Let us now consider the real problem, where the prediction is multimodal: Two image patches such as the ones shown in Fig. 2(c) can be similar but extracted from different training images and, therefore, correspond to different expected heatmaps. In other words, in our training set, we can have values for samples $i$, $i'$ and locations $(u, v)$ and $(u', v')$ such that $\mathcal{P}(I^{(i)}, u, v) \approx \mathcal{P}(I^{(i')}, u', v')$ and $\mathcal{H}^{(i)} \neq \mathcal{H}^{(i')}$.

It may seem as though, in this case, training given by Eq. (1) would fail or need to be modified. *In fact, Eq. (1) remains valid.* This is because we use the least-squares loss function: For image patches with similar appearances that correspond to multiple possible heatmaps, $f_\theta(\cdot)$ will learn to predict the average of these heatmaps, which is exactly what we want. The bottom row of Fig. 2 shows such heatmaps. At run-time, because we will combine the contribution of multiple image patches, we will be able to resolve the ambiguities.

## 4.2   Run-Time Inference

At run-time, given an input image $I$, we extract patches from randomly selected locations from the input image and feed them into the predictor $f_\theta(\cdot)$. To combine the contributions of the different patches, we use a simple ensemble approach and average the predicted heatmaps for each 2D projection. We take the locations of the global maxima after averaging them as the final predictions for the 2D projections.

More formally, the final prediction $\widetilde{\mathbf{p}_j}$ for the 2D projection $\mathbf{p}_j$ is the location of the global maximum of $\sum_{u,v} \text{Transl}(f_\theta(\mathcal{P}(I, u, v)), -u, -v)[j]$, the sum of the heatmaps predicted for the $j$-th projection, translated such that these heatmaps align correctly. The sum is performed over randomly sampled patches. An evaluation of the effect of the number of samples is provided in the supplementary material. To compute the pose, we use a P$n$P estimation with RANSAC [9] on the correspondences between the corners of the object's 3D bounding box and the $\widetilde{\mathbf{p}_j}$ locations.

### 4.3   Two-Step Procedure

In practice, we first estimate the 2D location of the object of interest, using the same method as in the previous subsection, but instead of sampling random locations, we apply the network $f_\theta(\cdot)$ in a sliding window fashion, as illustrated in Fig. 1 (b). For each image location, we compute a score by summing up the heatmap values over a bounding box of size 128×128 and over the 8 corners for each object, which is done efficiently using integral images. We apply Gaussian smoothing and thresholding to the resulting score map. We use the centers-of-mass of the regions after thresholding as the centers of the input image $I$. Finally, we use this image as input to the method described in the previous subsection. We use a fixed size for this region as our method is robust to scale changes.

## 5   Evaluation

In this section, we evaluate our method and compare it to the state-of-the-art. For this, we use two datasets: The Occluded LineMOD dataset [12], and the YCB-Video dataset [3]. Both datasets contain challenging sequences with partially occluded objects and cluttered backgrounds. In the following, we first provide the implementation details, the evaluation metrics used and then present the results of evaluation of the two datasets, including the results of an ablative analysis of our method.

### 5.1   Implementation Details

*Training Data:* The training data consist of real and synthetic images with annotated 3D poses and object masks, as was also the case in [3]. To render the synthetic objects, we use the models that are provided with the datasets. We crop the objects of interest from the training images and paste them onto random backgrounds [32] sampled from ImageNet [31] to achieve invariance to different backgrounds. We augment the dataset with small affine perturbations in HSV color space.

*Network Training:* The network is optimized using ADAM [33] with default parameters and using a minibatch size of 64, a learning rate of 0.001, and 100k iterations. We train one network per object starting from a random initialization.

*Symmetric Objects:* We adapt the heatmap generation to symmetric objects present in the two datasets. For rotationally symmetric objects, *e.g.*, cylindrical shapes, we only predict a single position around the rotation axis. For mirror-symmetric objects, we only train on half the range of the symmetry axis, as was performed in [5].

*Feature Mapping:* Optionally, we apply the Feature Mapping method as described in [11] to compensate for a lack of real training data. We apply the mapping between the FC1 and FC2 layers shown in Fig. 4. The mapping network uses the same architecture as described in [11], but the weight for the feature loss is significantly lower ($10^{-5}$).

### 5.2    Evaluation Metrics

We consider the most common metrics. The 2D Reprojection error [8] computes the distances between the projections of the 3D model points when projected using the ground truth pose, and when using the predicted pose. The ADD metric [34] calculates the average distance in 3D between the model points, after applying the ground truth pose and the predicted pose. For symmetric objects, the 3D distances are calculated between the closest 3D points, denoted as the ADI metric. Below, we refer to these two metrics as AD{D|I} and use the one appropriate to the object. The exact formulas for these metrics are provided in the supplementary material.

### 5.3    Occluded LineMOD Dataset

The Occluded LineMOD dataset [12] consists of a sequence of 1215 frames, each frame labeled with the 3D poses of 8 objects as well as object masks. The objects show severe occlusions, which makes pose estimation extremely challenging. The sequences were captured using an RGB-D camera with $640 \times 480px$ images, however, we use *only the color images* for our method and all results reported.

For training the heatmap predictors, we use the LineMOD dataset [34] that contains the same objects as the Occluded LineMOD dataset. This protocol is commonly used for the dataset [5, 7, 3, 4], since the Occluded LineMOD dataset only contains testing sequences. Fig. 5 shows some of the qualitative results obtained. We give an extensive quantitative evaluation in the following section.

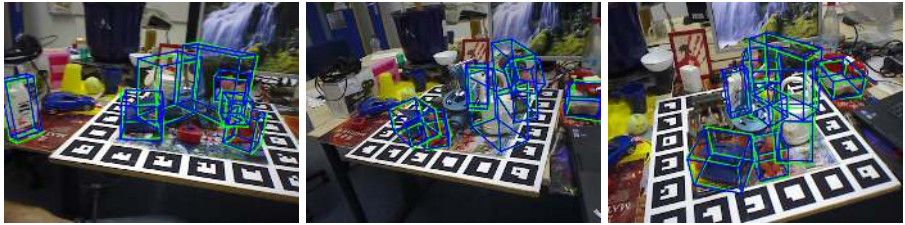

Fig. 5: Some qualitative results on the Occluded LineMOD dataset [12]. We show the 3D bounding boxes of the objects projected onto the color image. Ground truth poses are shown in green, and our predictions are shown in blue. More results are provided in the supplementary material.

**Quantitative Results** Fig. 6 shows the fraction of frames where the 2D Reprojection error is smaller than a given threshold, for each of the 8 objects from the dataset. A larger area under the curve denotes better results. We compare these results to those obtained from the use of several recent methods that also work only with color images, namely, BB8 [5], PoseCNN [3], Jafari *et al.* [4], and

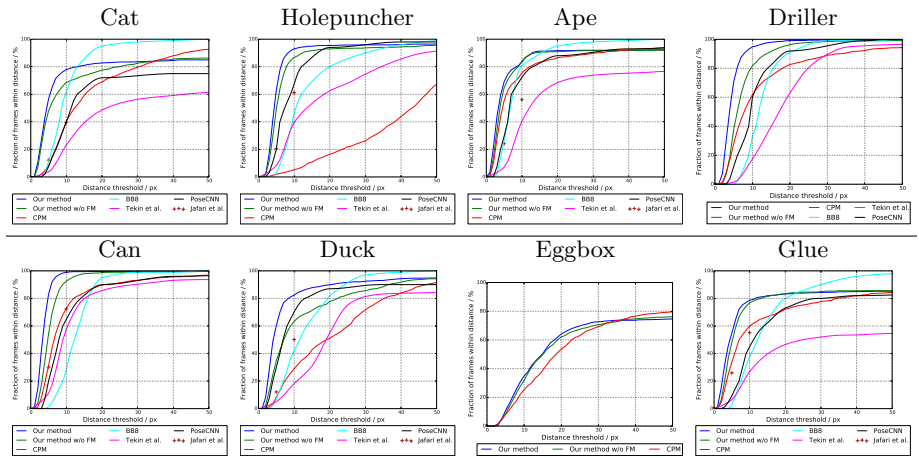

Fig. 6: Evaluation on the Occluded LineMOD dataset [12] using color images only. We plot the fraction of frames for which the 2D Reprojection error is smaller than the threshold on the horizontal axis. Our method provides results that significantly outperform those reported by previous work. "w/o FM" denotes without Feature Mapping.

Tekin *et al.* [7]. Note that the method described in [5] uses ground truth detection, whereas ours does not. Our method performs significantly more accurately on all sequences. Notably, we also provide results for the *Eggbox* object, which, so far, was not considered since it was too difficult to learn for [5, 7, 4].

Adding Feature Mapping [11] improves the 2D Reprojection error for a threshold of 5*px* by 17% on average. We also tried Feature Mapping for the approach of [5], but it did not improve the results because the occlusions influence the feature maps too greatly when the network input contains occluders, as already discussed in the introduction.

Further quantitative results are given in Table 1, where we provide the percentage of frames for which the ADD or ADI metric is smaller than 10% of the object diameter, as [3] reported such results on the Occluded LineMOD dataset. This is considered a highly challenging metric. We also give the percentage of frames that have a 2D Reprojection error of less than 5*px*. Our method significantly outperforms all other methods on these metrics by a large margin.

**The Effect of Seeing Occlusions During Training** We evaluate the importance of knowing the occluder in advance. [5, 7, 3] assumed that the occluder is another object from the LineMOD dataset and only used occlusions from these objects during training. However, in practice, this assumption does not hold, since the occluder can be an arbitrary object. Therefore, we investigated how the performance was affected by the use of occlusions during training.

Table 1: Comparison on the Occluded LineMOD dataset [12] with color images only. We provide the percentage of frames for which the AD{D|I} error is smaller than 10% of the object diameter, and for which the 2D Reprojection error is smaller than 5*px*. Objects marked with a * are considered to be symmetric.

| Method | AD{D\|I}-10% | | | | | | | | | 2D Reprojection Error-5px | | | | | | | | |
| --- | --- | --- | --- | --- | --- | --- | --- | --- | --- | --- | --- | --- | --- | --- | --- | --- | --- | --- |
| | Ape | Can | Cat | Driller | Duck | Eggbox* | Glue* | Holepun. | Average | Ape | Can | Cat | Driller | Duck | Eggbox* | Glue* | Holepun. | Average |
| PoseCNN [3] | 9.6 | *45.2* | 0.93 | 41.4 | **19.6** | 22.0 | 38.5 | **22.1** | 24.9 | 34.6 | 15.1 | 10.4 | 7.40 | 31.8 | 1.90 | 13.8 | 23.1 | 17.2 |
| Tekin *et al.* [7] | – | – | – | – | – | – | – | – | – | 7.01 | 11.2 | 3.62 | 1.40 | 5.07 | – | 6.53 | 8.26 | 6.16 |
| BB8 [5] | – | – | – | – | – | – | – | – | – | 28.5 | 1.20 | 9.60 | 0.00 | 6.80 | – | 4.70 | 2.40 | 7.60 |
| Jafari *et al.* [4] | – | – | – | – | – | – | – | – | – | 24.2 | 30.2 | 12.3 | – | 12.1 | – | 25.9 | 20.6 | 20.8 |
| CPM [27] | 12.5 | 25.6 | 1.43 | 23.8 | 6.99 | 18.3 | 15.0 | 0.74 | 13.0 | 55.4 | 30.6 | 15.7 | 27.9 | 26.6 | 7.97 | 18.5 | 1.81 | 23.1 |
| Our method w/o FM | *16.5* | *42.5* | *2.82* | *47.1* | *11.0* | *24.7* | *39.5* | *21.9* | *25.8* | *64.7* | *53.0* | *47.9* | *35.1* | *36.1* | *10.3* | *44.9* | *52.9* | *43.1* |
| Our method | **17.6** | **53.9** | **3.31** | **62.4** | *19.2* | **25.9** | **39.6** | 21.3 | **30.4** | **69.6** | **82.6** | **65.1** | **73.8** | **61.4** | **13.1** | **54.9** | **66.4** | **60.9** |

We compare our results (without Feature Mapping) to two state-of-the-art approaches: our reimplementations of BB8 [5] and CPM [27]. To avoid bias introduced by the limited amount of training data in the Occluded LineMOD dataset [12], we consider synthetic images both for training and for testing here.

We investigate three different training schemes: (a) No occlusions used for training; (b) random occlusions by simple geometric shapes; (c) random occlusions with the same objects from the dataset, as described in [5, 7, 4]. We compare the different training schemes in Fig. 7. Training without occlusions clearly result in worse performance for BB8 and CPM, whereas our method is significantly more robust. Adding random geometric occlusions during training slightly increases the performance of BB8 and CPM, since the networks learn invariance to occlusions, however, mainly for these specific occlusions, whereas our approach maintains the accuracy compared to training without occlusions. Using occluders from the dataset gives the best results, since the networks learn to ignore specific features from these occluders. This, however, is only possible when the occluders are known in advance, which is not necessarily the case in practice.

**Patch Size and Number of Patches** We evaluated the influence of the patch size on the predicted pose accuracy. There is a range of sizes (25px to 40px) for which the performances stay very close to those presented in Table 1. Small patches seem to lack discriminative power, the 2D Reprojection metric gets 19% worse with 8px patches, and large patches are sensitive to occlusions, which leads to a decrease in the 2D Reprojection metric of 5% for 128px patches.

In the supplementary material, we provide a detail study of the influence of the number of patches on the predicted pose accuracy. The main conclusions are that the accuracy starts to flatten when more than 64 patches are used, and that — if a preprocessing algorithm could be used to provide a segmentation mask — we could reduce the number of patches to achieve the same level of accuracy.

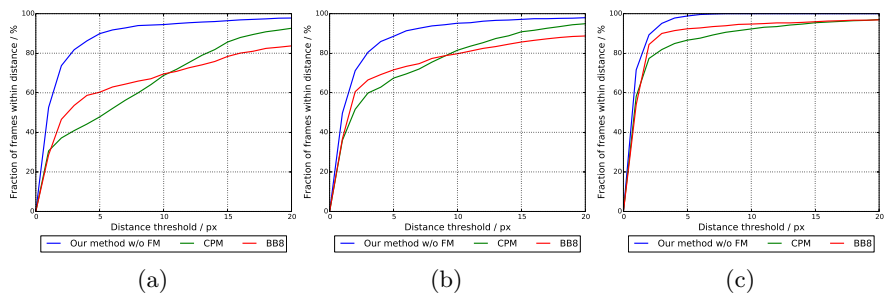

Fig. 7: Evaluation of synthetic renderings of scenes from the Occluded LineMOD dataset (see text) using the 2D Reprojection error. (a) Training without occlusions; (b) training with random geometric occlusions; and (c) training with occluding objects from the LineMOD dataset [34]. Knowing the occluders in advance significantly improves the performances of BB8 [5] and CPM [27], however, this knowledge is often not available in practice. Our method does not require this knowledge.

**Runtime** We implemented our method in Python on an Intel i7 with 3.2GHz and 64GB of RAM, using an nVidia GTX 980 Ti graphics card. Pose estimation is $100ms$ for 64 patches, and detection takes $150ms$ on a $640 \times 480$ camera frame. Predicting the heatmaps for a single patch takes $4ms$, and the total runtime could, thus, be significantly reduced by processing the individual patches in parallel.

### 5.4    YCB-Video Dataset

The recently proposed YCB-Video dataset [3] consists of 92 video sequences, where 12 sequences are used for testing and the remaining 80 sequences for training. In addition, the dataset contains 80k synthetically rendered images, which can be used for training as well. There are 21 objects in the dataset, which are taken from the YCB dataset [35] and are publicly available for purchase. The dataset is captured with two different RGB-D sensors, each providing $640 \times 480$ images, but we only use the color images. The test images are extremely challenging due to the presence of significant image noise and different illumination levels. Each image is annotated with the 3D object poses, as well as the objects' masks. Fig. 8 shows some qualitative results. We give an extensive quantitative evaluation in the following section.

**Quantitative Results** We provide the 2D Reprojection error and the AD{D|I} metrics averaged over all the objects in Table 2. In [3], the area under the accuracy-threshold curve was used as a metric, which we also provide.[3] Again, our approach results in better performance according to these metrics.

---

[3] The metrics are calculated from the results provided by the authors at their website.

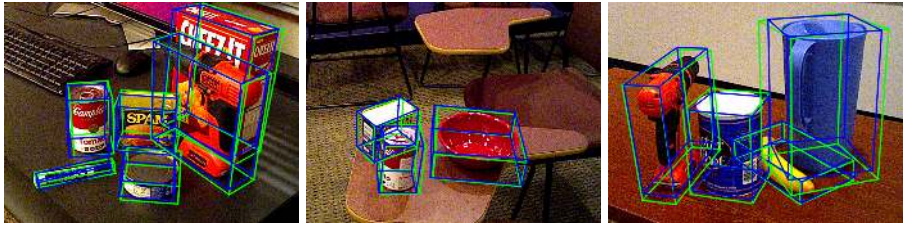

Fig. 8: Qualitative results on the YCB-Video dataset [3]. The green bounding boxes correspond to the ground truth poses, the blue ones to our estimated poses. More results are provided in the supplementary material.

Table 2: Comparison on the YCB-Video dataset [3]. We refer to the supplementary material for the object-specific numbers and additional plots. Our method clearly outperforms the baseline.

| Method | PoseCNN [3] | | | Our method w/o FM | | | Our method | | |
|---|---|---|---|---|---|---|---|---|---|
| | AUC | AD{D\|I}-10% | 2D Repr-5*px* | AUC | AD{D\|I}-10% | 2D Repr-5*px* | AUC | AD{D\|I}-10% | 2D Repr-5*px* |
| Average | 61.0 | 21.3 | 3.72 | *61.4* | *33.6* | *23.1* | **72.8** | **53.1** | **39.4** |

## 6   Discussion and Conclusion

In this paper, we introduced a novel method for 3D object pose estimation that is inherently robust to partial occlusions of the object. To do this, we considered only small image patches as input and merged their contributions. Because we chose to compute the pose by initially predicting the 2D projections of 3D points related to the object, the prediction can be performed in the form of 2D heatmaps. Since heatmaps are closely related to density functions, they can be conveniently applied to capture the ambiguities that arise when using small image patches as input. We showed that training a network to predict the heatmaps in the presence of such ambiguities is much simpler than it may sound. This resulted in a simple pipeline, which outperformed much more complex methods on two challenging datasets.

Our approach can be extended in different ways. The heatmaps could be merged in a way that is more robust to erroneous values than simple averaging. The pose could be estimated by considering the best local maxima rather than only the global maxima. Sampling only patches intersecting with the object mask, which could be predicted by a segmentation method, would limit the influence of occluders and background in the accumulated heatmaps even more. Predicting the heatmaps could be performed in parallel.

**Acknowledgment** This work was funded by the Christian Doppler Laboratory for Semantic 3D Computer Vision. We would like to thank Yu Xiang for providing additional results.

# Supplementary Material:
# Making Deep Heatmaps Robust to Partial Occlusions for 3D Object Pose Estimation

Markus Oberweger[1][0000−0003−4247−2818], Mahdi Rad[1][0000−0002−4011−4729], and Vincent Lepetit[2,1][0000−0001−9985−4433]

[1] Institute for Computer Graphics and Vision, Graz University of Technology, Graz, Austria
[2] Laboratoire Bordelais de Recherche en Informatique, Université de Bordeaux, Bordeaux, France
{oberweger,rad,lepetit}@icg.tugraz.at

**Abstract.** We present the supplementary material. Specifically, we state the formulas for the evaluation metric, show visualizations of the occluded feature maps, report more details on the quantitative evaluation on the YCB-Video dataset, and finally give additional qualitative results on the Occluded LineMOD dataset [1] and the YCB-Video dataset [2].

## 1 Supplementary Material

### 1.1 Evaluation Metrics

For the evaluation of the 3D object pose, we apply the most common metrics [3, 4, 2]: Firstly, the 2D Reprojection error. This error computes the distances between the projected 3D model points $\mathcal{M}$ in 2D. Hereby, $\mathbf{Rt}$ denotes the ground truth pose, $\widehat{\mathbf{Rt}}$ denotes our estimated pose, and $\mathbf{K}$ denotes the intrinsic camera calibration matrix. The 2D Reprojection error can be calculated as:

$$\frac{1}{|\mathcal{M}|} \sum_{\mathbf{x} \in \mathcal{M}} \|\mathbf{K} \cdot \mathbf{Rt} \cdot \mathbf{x} - \mathbf{K} \cdot \widehat{\mathbf{Rt}} \cdot \mathbf{x}\| \tag{1}$$

Similarly, the ADD metric calculates the average distance in 3D between the model points transformed by ground truth pose and our estimated pose:

$$\text{ADD} = \frac{1}{|\mathcal{M}|} \sum_{\mathbf{x} \in \mathcal{M}} \|\mathbf{Rt} \cdot \mathbf{x} - \widehat{\mathbf{Rt}} \cdot \mathbf{x}\| \tag{2}$$

Further, for symmetric objects, the 3D distances are calculated between the closest 3D points, referred to as the ADI metric:

$$\text{ADI} = \frac{1}{|\mathcal{M}|} \sum_{\mathbf{x}_1 \in \mathcal{M}} \min_{\mathbf{x}_2 \in \mathcal{M}} \|\mathbf{Rt} \cdot \mathbf{x}_1 - \widehat{\mathbf{Rt}} \cdot \mathbf{x}_2\| \tag{3}$$

## 1.2   Visualization of Occluded Feature Maps

Fig. 1 shows the effect of occlusions on the feature maps of CNNs. We show two examples: in the top part of the figure an example with small occlusion and in the bottom part of the figure an example with a larger occlusion. In each part of the figure, we show the feature maps with and without occlusions together with the sums of the squared differences between them. Further, we compare when the network is trained with or without occlusion examples. The left column shows the results for a feedforward network [5] and the right column shows the same for a Convolutional Pose Machine [6]. The influence of the occlusion increases with the layers' depths, as receptive fields are larger in the deeper layers than in the first layers, even when the method is trained with occlusion examples.

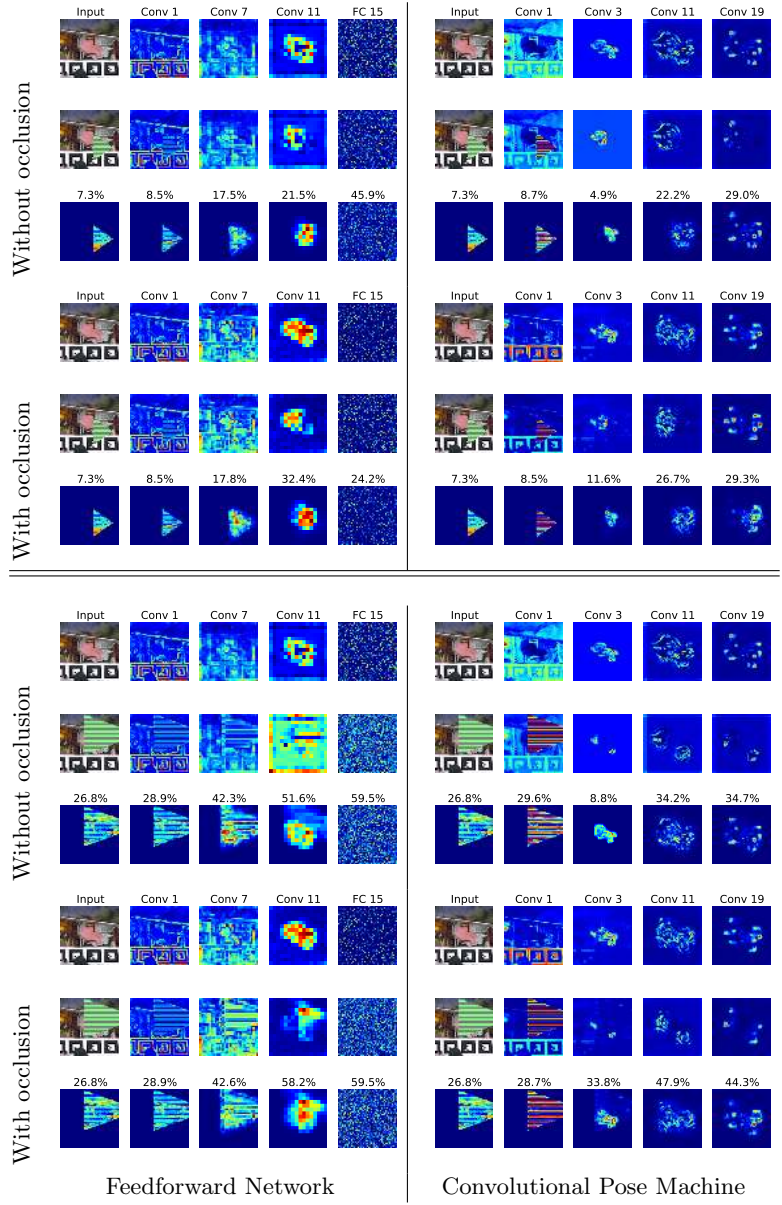

Fig. 1: Effect of occlusions on the feature maps of CNNs. See text for details.

### 1.3   Number of Patches

Table 1 shows the aggregated heatmaps for one input image when varying the number of patches.

Table 1: Effect of the number of patches used at run-time. While for small numbers of patches, the resulting heatmaps are strongly effected by the sampling, the aggregation of more heatmaps forms more accurate distributions robust to the patches sampled from occluded regions.

| # Patches | Heatmaps |
| --- | --- |
| GT | 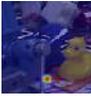 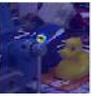 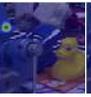 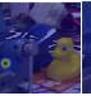 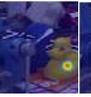 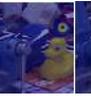 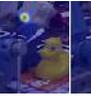 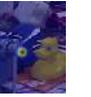 |
| $N = 1$ | 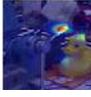 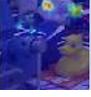 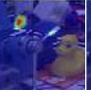 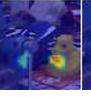 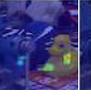 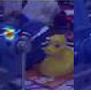 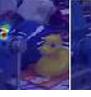 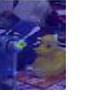 |
| $N = 16$ | 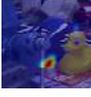 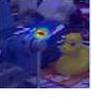 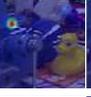 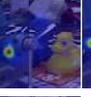 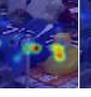 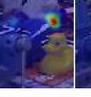 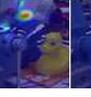 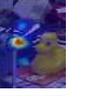 |
| $N = 32$ | 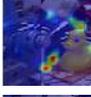 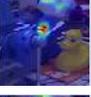 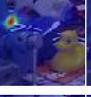 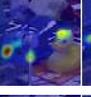 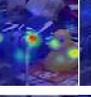 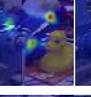 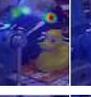 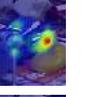 |
| $N = 64$ | 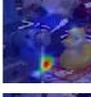 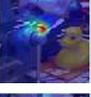 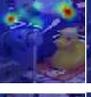 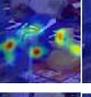 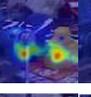 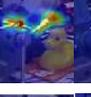 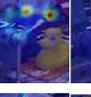 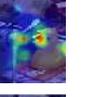 |
| $N = 512$ | 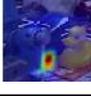 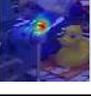 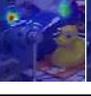 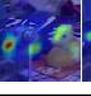 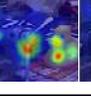 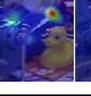 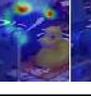 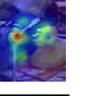 |

Fig. 2(a) shows the influence of the sampling of the patches on the 2D Reprojection error for different sampling strategies. Since we do not have a segmentation mask of the object given during runtime, we rely on random sampling. However, if we sample enough patches, *i.e.* $> 64$, we can achieve the same performance as sampling the patches only on the object mask that we created for the test sequences. So, if some preprocessing algorithm would provide a segmentation mask as well, we could reduce the number of patches to achieve the same accuracy. Fig. 2(b) shows the influence of the number of patches that we randomly sample. The more patches we sample, the more accurate the results get, however, the accuracy starts to flatten for $> 64$ patches.

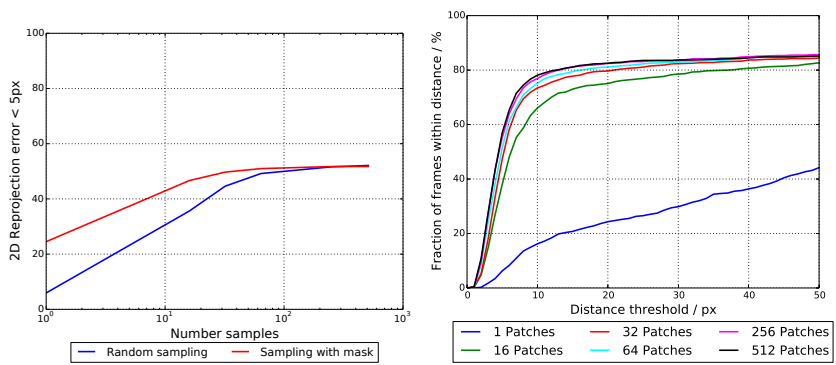

(a) Sampling patches within the object mask versus random sampling.

(b) Influence of number of samples.

Fig. 2: (a) Evaluation of the sampling strategy and (b) the number of sampled patches on the Occluded LineMOD dataset [1] for the *Cat* object. We plot the fraction of frames where the 2D Reprojection error is smaller than a threshold.

### 1.4 Quantitative Evaluation YCB-Video

A detailed evaluation of the different objects of the YCB-Video dataset is shown in Table 2. Our method performs better for almost all objects, and significantly better on all metrics on average.

Table 2: Comparison on the YCB-Video dataset [2]. We denote the area under the accuracy-threshold curve (AUC), the percentage of frames which have a 2D Reprojection error of less than 5*px*, and the percentage of frames where the 3D ADD or ADI error is less than 10% of the object diameter. The objects marked with * are considered to be symmetric.

| Method / Object | PoseCNN [2] | | | Our method w/o FM | | | Our method | | |
|---|---|---|---|---|---|---|---|---|---|
| | AUC | AD{D\|I}-10% | 2D Repr-5px | AUC | AD{D\|I}-10% | 2D Repr-5px | AUC | AD{D\|I}-10% | 2D Repr-5px |
| 002_master_chef_can | 50.1 | 3.58 | 0.09 | 68.5 | 32.9 | 9.94 | **81.6** | **75.8** | **29.7** |
| 003_cracker_box | 52.9 | 25.1 | 0.12 | 74.7 | 62.6 | 24.5 | **83.6** | **86.2** | **64.7** |
| 004_sugar_box | 68.3 | 40.3 | 7.11 | 74.9 | 44.5 | 47.0 | **82.0** | **67.7** | **72.2** |
| 005_tomato_soup_can | 66.1 | 25.5 | 5.21 | 68.7 | 31.1 | **41.5** | **79.7** | **38.1** | *39.8* |
| 006_mustard_bottle | 80.8 | *61.9* | 6.44 | 72.6 | 42.0 | *42.3* | **91.4** | **95.2** | **87.7** |
| 007_tuna_fish_can | **70.6** | **11.4** | 2.96 | 38.2 | *6.79* | *7.14* | *49.2* | 5.83 | **38.9** |
| 008_pudding_box | 62.2 | 14.5 | 5.14 | 82.9 | 58.4 | 43.9 | **90.1** | **82.2** | **78.0** |
| 009_gelatin_box | 74.8 | 12.1 | 15.8 | 82.8 | 42.5 | 62.1 | **93.6** | **87.8** | **94.8** |
| 010_potted_meat_can | 59.5 | 18.9 | 23.1 | 66.8 | 37.6 | 38.5 | **79.0** | **46.5** | **41.2** |
| 011_banana | **72.1** | *30.3* | 0.26 | 44.9 | 16.8 | *8.18* | *51.9* | **30.8** | **10.3** |
| 019_pitcher_base | 53.1 | 15.6 | 0.00 | **70.3** | 57.2 | **15.9** | 69.4 | **57.9** | *5.43* |
| 021_bleach_cleanser | 50.2 | 21.2 | 1.16 | 67.1 | 65.3 | 12.1 | **76.1** | **73.3** | **23.2** |
| 024_bowl* | *69.8* | 12.1 | 4.43 | 58.6 | 25.6 | 16.0 | **76.9** | **36.9** | **26.1** |
| 025_mug | **58.4** | 5.18 | 0.78 | 38.0 | *11.6* | 20.3 | *53.7* | **17.5** | **29.2** |
| 035_power_drill | 55.2 | 29.9 | 3.31 | 72.6 | 46.1 | 40.9 | **82.7** | **78.8** | **69.5** |
| 036_wood_block* | **61.8** | 10.7 | 0.00 | 57.7 | **34.3** | **2.48** | 55.0 | *33.9* | *2.06* |
| 037_scissors | *35.3* | *2.21* | 0.00 | 30.9 | 0.00 | 0.00 | **65.9** | **43.1** | **12.1** |
| 040_large_marker | **58.1** | *3.39* | *1.38* | 46.2 | 3.24 | 0.00 | *56.4* | **8.88** | **1.85** |
| 051_large_clamp* | *50.1* | 28.5 | *0.28* | 42.4 | 10.8 | 0.00 | **67.5** | **50.1** | **24.2** |
| 052_extra_large_clamp* | 46.5 | 19.6 | *0.58* | *48.1* | 29.6 | 0.00 | **53.9** | **32.5** | **1.32** |
| 061_foam_brick* | *85.9* | *54.5* | 0.00 | 82.7 | 51.7 | *52.4* | **89.0** | **66.3** | **75.0** |
| Average | 61.0 | 21.3 | 3.72 | *61.4* | *33.6* | *23.1* | **72.8** | **53.1** | **39.4** |

Fig. 3(a) and (b) plot the fraction of frames where the 2D Reprojection error and the AD{D|I} metrics averaged over all the objects are smaller than a given threshold. Our approach clearly results in better performance.

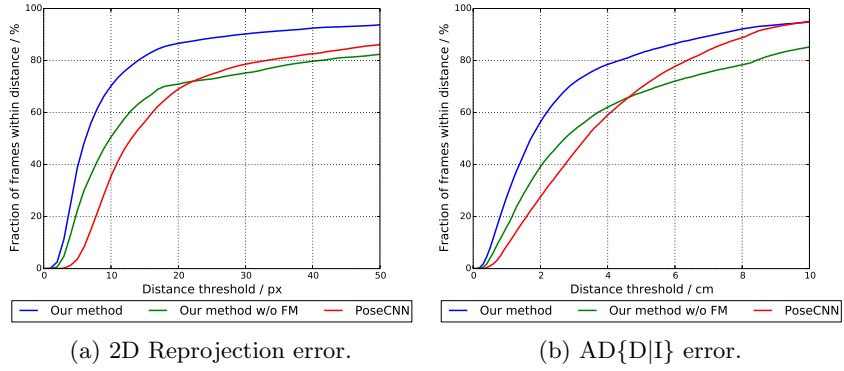

(a) 2D Reprojection error.    (b) AD{D|I} error.

Fig. 3: Evaluation on the YCB-Video dataset [2]. We plot the fraction of frames where (a) the 2D Reprojection error and (b) AD{D|I} metrics are smaller than a threshold.

### 1.5   Qualitative Results Occluded LineMOD

We show qualitative results on the Occluded LineMOD dataset [1] in Fig. 4.

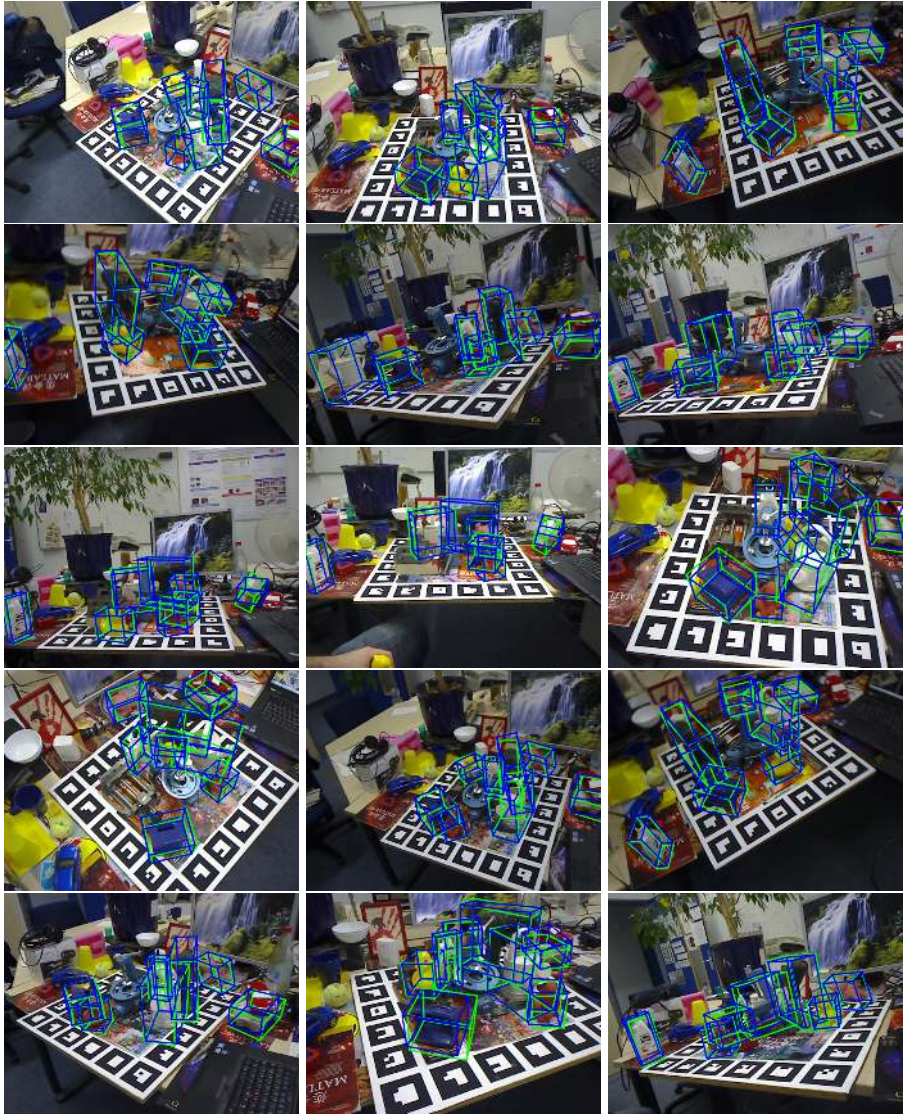

Fig. 4: Qualitative results on the Occluded LineMOD dataset [1]. We show the 3D bounding boxes of the objects projected to the color image. Ground truth poses are shown in green, our predictions are shown in blue.

### 1.6   Qualitative Results YCB-Video

We show qualitative results on the YCB-Video dataset [2] in Fig. 5. Further, we would like to refer to the supplementary video for more results on the YCB-Video dataset.

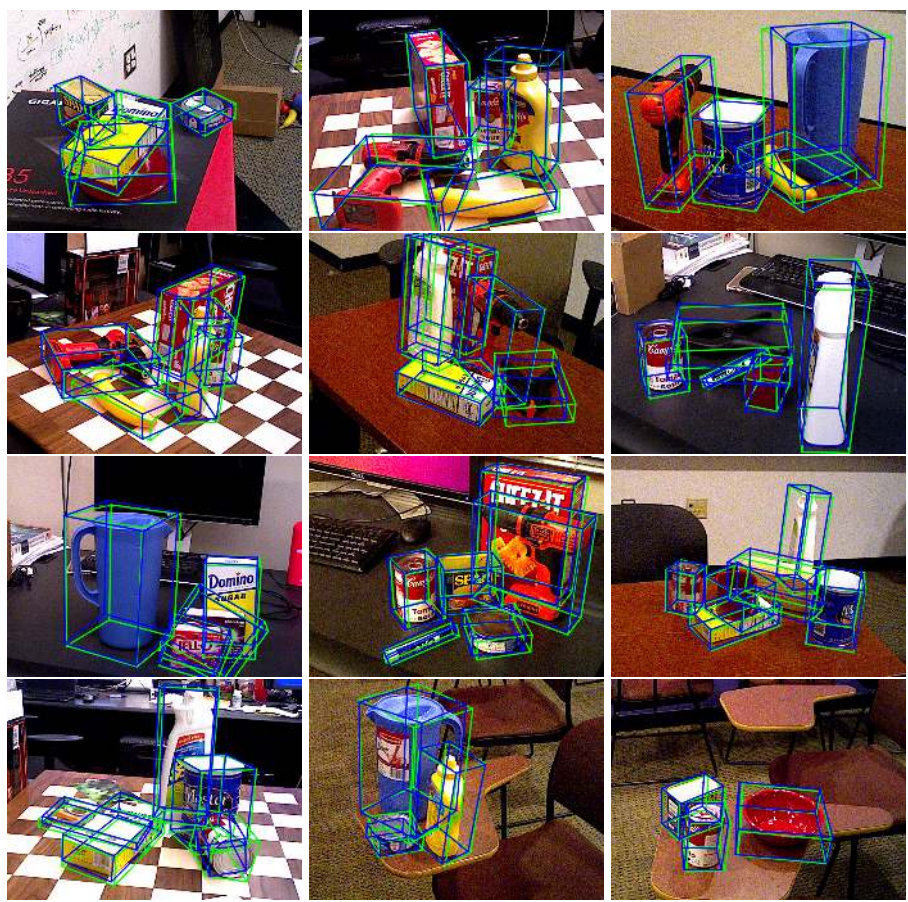

Fig. 5: Qualitative results on the YCB-Video dataset [2]. We show the 3D bounding boxes of the objects projected to the color image. Ground truth poses are shown in green, our predictions are shown in blue.